\documentclass[runningheads]{llncs}
\usepackage{bm}
\usepackage{multirow}
\usepackage{graphicx}
\usepackage{subfigure}
\usepackage{amsmath}
\usepackage{amssymb}
\usepackage{graphicx}
\usepackage[linkcolor=green]{hyperref}
\usepackage{url}
\usepackage{color}
\usepackage{threeparttable}
\usepackage{fixltx2e}

\newenvironment{sequation}{\begin{equation}\small}{\end{equation}}
\newenvironment{sequation*}{\begin{equation*}\small}{\end{equation*}}

\MakeRobust{\overleftrightarrow}
\begin{document}
\title{Label-aware Document Representation via Hybrid Attention for Extreme Multi-Label Text Classification}
%
\titlerunning{Label-aware Document Representation via Hybrid Attention}
%
\author{Xin Huang \and
Boli Chen \and
Lin Xiao \and
Liping Jing}
\authorrunning{Huang et al.}

%
\institute{Beijing Key Lab of Traffic Data Analysis and Mining, Beijing Jiaotong University, Beijing 100044, China
\\\email{\{18120367, 18120345, 17112079, lpjing\}@bjtu.edu.cn}}
\maketitle              
\begin{abstract}Extreme multi-label text classification (XMTC) aims at tagging a document with most relevant labels from an extremely large-scale label set. It is a challenging problem especially for the tail labels because there are only few training documents to build classifier. This paper is motivated to better explore the semantic relationship between each document and extreme labels by taking advantage of both document content and label correlation. Our objective is to establish an explicit \textbf{l}abel-\textbf{a}ware representation for each document with a \textbf{h}ybrid \textbf{a}ttention deep neural network model(\textit{LAHA}). \textit{LAHA} consists of three parts. The first part adopts a multi-label self-attention mechanism to detect the contribution of each word to labels. The second part exploits the label structure and document content to determine the semantic connection between words and labels in a same latent space. An adaptive fusion strategy is designed in the third part to obtain the final label-aware document representation so that the essence of previous two parts can be sufficiently integrated. Extensive experiments have been conducted on five benchmark datasets by comparing with the state-of-the-art methods. The results show the superiority of our proposed \textit{LAHA} method, especially on the tail labels.
	
\end{abstract}
\vspace{-2mm}
\section{Introduction}
Extreme multi-label text classification(XMTC) aims at automatically tagging a document with most relevant labels from an extremely large label set. For instance, there are millions of categories on Wikipedia and one might wish to build a classifier that can annotate a given message with the subset of most relevant categories~\cite{ref_article9}. XMTC has become increasingly important due to the boom of big data, while it becomes significantly challenging because it has to simultaneously handle massive documents, features and labels. Thus it is emergency to develop effective extreme multi-label classifer for various real-applications such as product categorization in e-commerce, news annotation and etc.

Multi-label text classification, unlike the traditional multi-class classification, allows for the co-existence of more than one labels for a single document. Meanwhile, there may be a large number of 'tail labels' with very few positive documents in XMTC tasks. To tackle the aforementioned issues, researchers pay much attention on two facets: 1) how to represent label so that the correlation among labels can be accurately mined, and 2) how to represent document so that the dependency among text can be sufficiently captured. Recently, state-of-the-art extreme multi-label learning methods have been proposed in each facet. 
Among them, tree-based and embedding-based methods become popular to find the label correlation as they can obtain notable accuracy improvement by constructing a hierarchy structure \cite{ref_article18} or learning a low-dimensional latent space \cite{ref_article9}.
Deep learning-based methods (e.g., convolutional neural network \cite{ref_article6}) have achieved great success to represent text data. These methods usually characterize one document with the same representation on all labels. In this case, the probability of document belonging to a class is determined by their overall matching score regardless of the label-aware semantic information. Recent works, \textit{AttentionXML}~\cite{ref_article7} and \textit{EXAM}~\cite{ref_article8}, turn attention to this issue with the aid of attentive neural network. However, they only focus on document or label content but ignoring the label structure among extreme labels which has been proved very important in extreme multi-label learning~\cite{ref_article9}.


To solve the above-mentioned problems, we introduce a \textbf{L}abel-\textbf{A}ware document representation model via a \textbf{H}ybrid \textbf{A}ttention neural network (\textit{LAHA}) by considering both document content and label structure. \textit{LAHA} consists of three parts. The first part aims at detecting the importance of each word to all labels via a self-attention bidirectional LSTM neural network. The second part tries to explore the semantic connection between words and labels in a latent space. Here the word embedding is obtained by the bidirectional LSTM neural network. The label embedding is determined from the label co-exist graph so that the label structure can be sufficiently maintained in the same latent space with words'. Based on these two embeddings, we introduce an interaction-attention mechanism to explicitly compute the semantic relation between the words and labels. The last part is to represent each document along each label via an adaptive fusion strategy. The goal of fusion strategy is to adaptively extract proper information from the previous two parts so that the final document representation has discriminative ability to construct classifier. 

 The proposed XMTC model \textit{LAHA} has been evaluated on five benchmark datasets and get competitive results, we summarize the major contributions.
	\begin{itemize}
	\item \textit{LAHA} is the first work to construct label-aware document representation by simultaneously considering document content and label structure. 
	\item The hybrid attention mechanism is firstly designed to adaptively extract the semantic relation between each document and all labels for XMTC.
	\item The performance of \textit{LAHA} was thoroughly investigated on widely-used benchmark datasets, indicating the advantage over the baselines.
	\item The code and hyper-parameter settings are released\footnote{\url{https://github.com/HX-idiot/Hybrid_Attention_XML}} to facilitate other researchers.
\end{itemize}
\vspace{-2mm}
\section{Related Work}
Significant progress has been made for XMTC. They can be roughly categorized into two categories: embedding-based and tree-based methods. Recently, due to the powerful ability of representation, deep learning technology has been introduced to effectively represent document for XMTC tasks. Next, we will briefly review them.
\vspace{-2mm}
\subsection{Embedding-based Methods}
Embedding-based methods aim at reducing the huge label space to a low dimensional space while preserving the label correlation as much as possible, and then compressed label embedding are decompressed for prediction. Various approaches have been presented such as compressed sensing~\cite{ref_article19}, output codes~\cite{ref_article20}, Singular Value Decomposition~\cite{ref_article21}, landmark labels~\cite{ref_article22}, Bloom filters~\cite{ref_article23}, etc. To efficiently handle large-scale label set, these embedding-based methods usually assume that the label matrix is low-rank. However, such methods have been proved unable to deliver high prediction accuracies as the low rank assumption is violated in most real world applications~\cite{ref_article9}. \textit{SLEEC}~\cite{ref_article9} can be taken as the most representative embedding-based method due to its significant accuracy and computationally efficiency. Its main idea is to learn a small ensemble of local distance preserving embeddings. Specifically, \textit{SLEEC} divides the training data set into several clusters, and in each cluster it detects embedding vectors by capturing non-linear label correlation and preserving the pairwise distance between labels. The $k$-nearest neighbors search is used to do prediction only in the cluster into which the test document is fallen. Later, Zhang et al.~\cite{ref_article15} adopted deep neural network for non-linear modeling the label embedding. Although these methods perform well, they play a heavy price in terms of prediction accuracy due to the loss of information during the compression and decompression phases.

\vspace{-2mm}
\subsection{Tree-based Methods}
Tree-based methods introduce a tree structure to divide the documents recursively at each non-leaf node, so that documents in each leaf node share similar label distribution. The most representative method \textit{FastXML}~\cite{ref_article18} implements this process by optimizing the normalized discounted cumulative gain (nDCG)-based ranking loss function.
Then, a base classifier is trained at each leaf node which only focuses on a few active labels. To enhance the robustness of predictions, an ensemble of multiple induced trees are learned. The main advantage of tree-based methods is that the prediction time complexity is typically sublinear in the training-set size and would be logarithmic if the induced tree is balanced. 

A recent extension work of \textit{FastXML} is \textit{PfastreXML}~\cite{ref_article10}, which adopted a propensity scored objective function instead of nDCG-based loss which is more friendly to tail labels.  \textit{Parabel}~\cite{ref_article11} is another tree-based method, which constructs balanced trees partitioning labels rather than instances. These tree-based methods represent document via bag-of-words,  where the words are treated as independent features, which will ignore the semantic dependency among words.
\vspace{-2mm}
\subsection{Deep Learning-based Methods}
To capture semantic dependency among words, researchers adopted deep learning models in text classification task due to its strong ability of representation. 
The popular deep models include CNN~\cite{ref_article27}, GRU~\cite{ref_article26}, RNN~\cite{ref_article13}, LSTM~\cite{ref_article14}, Bi-LSTM~\cite{ref_article28}, BERT~\cite{ref_article30} and several combination networks~\cite{ref_article24,ref_article29}. Even though they have achieved great success in traditional NLP tasks, few work is designed for XMTC.

\textit{XML-CNN}~\cite{ref_article6} can be taken as the first and most representative work using deep learning model in XMTC. It takes advantage of CNN, dynamic max-pooling and bottle-neck layer to build the deep model. Due to the limited window size, \textit{XML-CNN} can not capture the long-distance dependency among text. Later, GRU and Bi-LSTM language models are adopted in \textit{AttentionXML}~\cite{ref_article7} and \textit{EXAM}~\cite{ref_article8} to effectively represent document for XMTC. Meanwhile, these two methods consider the difference of one document represesntation along different labels. \textit{AttentionXML}~\cite{ref_article7} adopts self-attention mechanism~\cite{ref_article16}, while \textit{EXAM}~\cite{ref_article8} exploits the label content information to calculate the relations between words and classes.
Although \textit{AttentionXML} obtains promising performance, it ignores the label structure which has been proved very important in embedding-based and tree-based multi-label learning methods. 

Therefore, in this paper, we propose a new XMTC deep model with hybrid attention to build label-aware document representation, which sufficiently exploits both document content and label structure.

\vspace{-4mm}
\section{LAHA model}

\begin{figure}[!h]
	\centering
	\includegraphics[width=0.9\textwidth, height=0.4\textwidth]{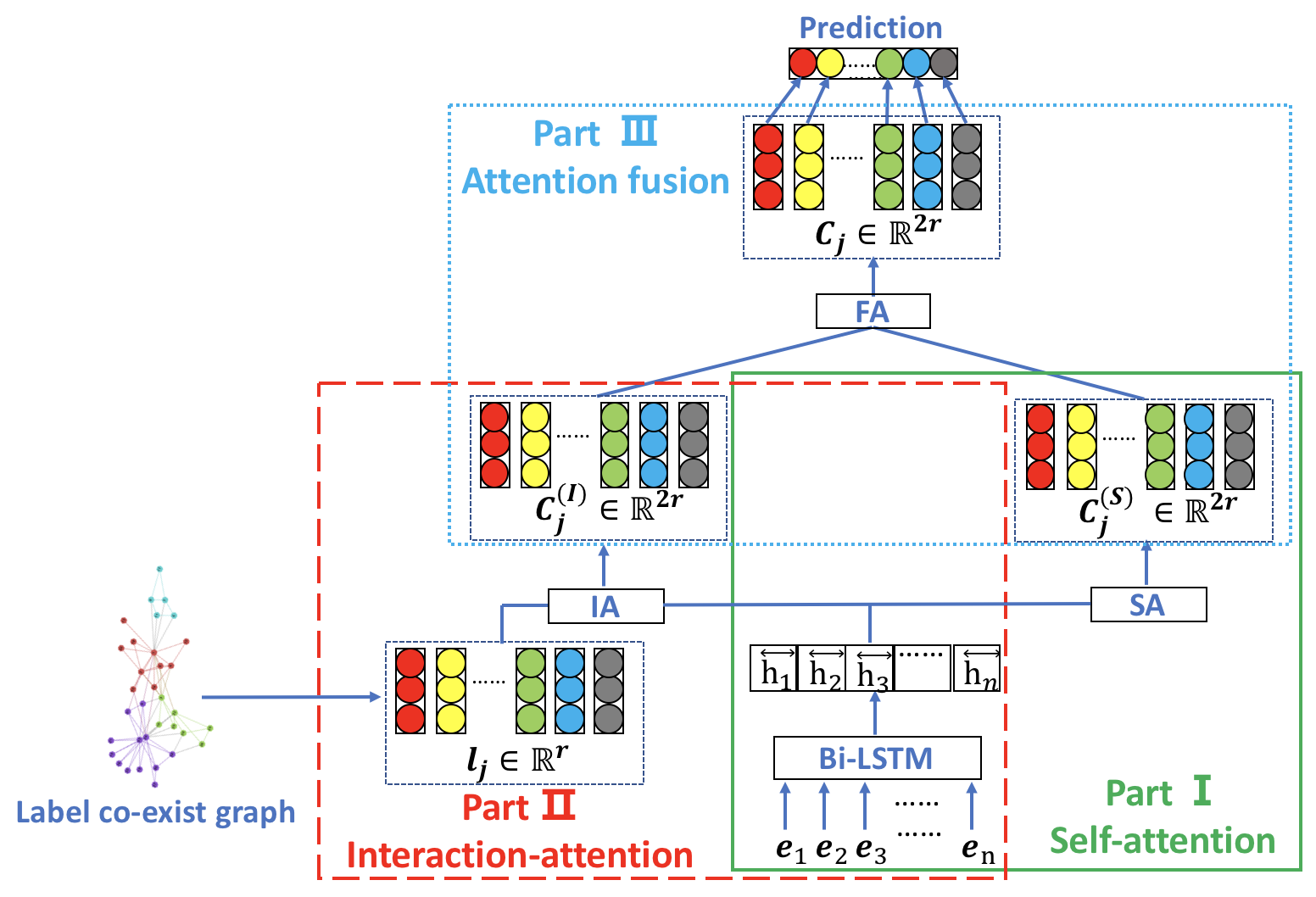}	
	\setlength{\abovecaptionskip}{0.cm}
	\setlength{\belowcaptionskip}{-0.cm}
	\caption{The architecture of \textbf{LAHA}. The solid green box indicates the self-attention process, the dashed red box represents interaction-attention process, and the dotted blue box indicates attention fusion to integrate self-attention and interaction-attention.} \label{Fig:LAHA}
\end{figure}

In this section, we introduce the proposed deep model (\textit{LAHA}) to handle XMTC tasks. The overall structure of \textit{LAHA} is shown in Fig.~\ref{Fig:LAHA}).
Our goal is to build a multi-label learning model from the training documents with a large-size label set. 
Let $D={ \{ \mathbf{({ x }_{ 1 },{ y }_{ 1 }),...,({ x }_{N},{ y }_{N})}\}  }$ be the given raw training document set containing total $N$ documents and belonging to $k$ labels. Each document has $n$ tokens (or words) and each word is represented via a $d$-dimensional deep semantic dense vector acquired from word2vec technique, $\mathbf{{ e }_{ t }}\in { \mathbb{R} }^{ d }$ ($t=1, ..., n$). $\mathbf{y_{i}}\subseteq \{ 0,1\} ^ { k }$ is the corresponding label vector, and $y_{ij }=1$ iff the $j$-th label is turned on for the $i$-th document $\mathbf{{x}_{i}}=(\mathbf{{ e }_1,...,e_n})$. 


\vspace{-2mm}
\subsection{Feature Embedding}

To build the proposed \textit{LAHA} multi-label text classifier, the raw text data is preprocessed via word embedding technique so that each word is represented as a low-dimensional dense vector. The extreme labels are embedded into dense vectors from the label co-exist graph so that the label correlation and local structure can be sufficiently captured.
%
\vspace{-2mm}
\subsubsection{Word Embedding} 

Once having the $d$-dimensional word vector $\mathbf{{ e }_{ t }}\in { \mathbb{R} }^{ d }$ for each word ($t=1, ..., n$), the whole document can be taken as a sequence of words ($\mathbf{{ e }_1,...,e_n}$) as the input of \textit{LAHA}. In order to capture the bi-directional contextual information, we adopt Bi-LSTM~\cite{ref_article28} to learn the word embedding for each input document. So the whole output of Bi-LSTM can be obtained by 
\begin{sequation}
H = ({ H }^{ (f) };{ H }^{ (b) })  \text{ with } 
{ H }^{ (f) }=(\overrightarrow { { h }_{ 1 } } ,...,\overrightarrow { { h }_{ n } } ) \in { \mathbb{R} }^{ r \times n} ;\quad
{ H }^{ (b) }=(\overleftarrow { { h }_{ 1 } } ,...,\overleftarrow { { h }_{ n } } ) \in { \mathbb{R} }^{ r \times n}
\end{sequation}
where $\mathbf{\overrightarrow { { h }_{ t } }} \in { \mathbb{R} }^{ r }$ and $\mathbf{\overleftarrow { { h }_{ t } }} \in { \mathbb{R} }^{ r }$ are the forward and backward word context representations respectively.
The whole document is taken as a matrix $H\in { \mathbb{R} }^{ 2r \times n}$.


\vspace{-2mm}
\subsubsection{Label Embedding}


To better extract label correlation information, we firstly build a label co-exist graph from the training data where each labels are represented by nodes. There will be an edge connecting the $i$-th label and the $j$-th label if they share at least one document~\cite{ref_article15}. Our goal is to represent the extreme labels in a low-dimensional latent space so that two nearby labels in the graph have similar representation, i.e., the local structure among labels are preserved as much as possible. Thus, the popular and powerful node2vec~\cite{ref_article17} is adopted here because it has ability to explore the labels' diverse neighborhoods by a flexible biased random walk procedure in a breadth-first sampling as well as depth-first sampling fashion. Finally, each label will be represented by a $r$-dimensional dense vector, i.e., $\mathbf{{ l }_{ i }}\in { \mathbb{R} }^{ r }$ for the i-th label ($i=1, ..., k$) and the whole label set can be described by $L=(\mathbf{{ l }_{ 1 },{ l }_{ 2 }, ..., { l }_{ k }}) \in \mathbb{R}^{r\times k}$.

\vspace{-4mm}
\subsection{Hybrid Attention Mechanism}
Hybrid attention mechanism aims at better representing each document by taking advantage of both document content and label structure. It is composed of self-attention mechanism on document content and interaction-attention mechanism to exploit document content and label structure.
\vspace{-4mm}
\subsubsection{Self-attention (SA)} has been successful used in text mining tasks such as relation extraction~\cite{ref_article31}. In multi-label data, since one document may be tagged by more than one labels, each document should have the most relative context to its corresponding labels. That is, the words in one document make different contributions to each label. To focus on different aspects of document, thus, we introduce self-attention mechanism (SA) \cite{ref_article16} on the output of Bi-LSTM ($H$). The attention score $A^{(S)}\in { \mathbb{R} }^{ n\times k}$ is calculated by 
\begin{sequation}
T=tanh({W}_{s1}H); \quad {A}^{(S)} =softmax({W}_{s2}T)
\end{sequation}
where ${ W }_{ { s }_{ 1 } }\in { \mathbb{R} }^{ { d }_{ a }\times 2r }$ and ${ W }_{ { s }_{ 2 } }\in { \mathbb{R} }^{ k\times { d }_{ a } }$ are parameters to be trained. 
$A^{(S)}_j\in{ \mathbb{R} }^{ n }$ is the attention scores of words along the $j$-th label. To efficiently handle extreme multi-label data, we adopt negative sampling strategy~\cite{ref_article12} to update ${ W }_{ { s }_{ 2 } }$ and computer $A^{(S)}_j$, so that all positive labels and a random small subset of negative labels are considered. 
Then, we can obtain the linear combination of context words for each label through self-attention mechanism as ${C}^{(S)}_j=H A^{(S)}_j$, which can be taken as the representation of the input document along the $j$-th lable. The whole matrix $C^{(S)}\in \mathbb{R}^{2r\times k}$ is the label-aware document represenation under the self-attention mechanism.


%
\vspace{-2mm}
\subsubsection{Interaction-attention (IA)} aims to determine the semantic connection between words and labels in a latent space. With the help of word embedding and label embedding technique, all words and labels are represented in the $r$-dimensional latent space as $H=(H^{ (f) };H^{ (b) })$ and $L$ respectively. To conveniently align the latent space of words and that of labels, a bridge mapping marix $W_q \in \mathbb{R}^{r\times r}$ is trained via $Q={ W }_{ q}{L}$. Similar to SA, we can do negative sampling on $L$ to produce $L^{*}\in{\mathbb{R} }^{ r\times { k^{ * } }}$ that is extracted from $L$ according to sampled indices, and just use $L^{*}$ for the following computation.

 Inspired by the interaction mechanism~\cite{ref_article8}, we take $Q\in \mathbb{R}^{r\times k}$ as the attention querys for each label, and use $H$ to construct the key-value pairs in terms of forward and backward information for each word. Then, the interactive matching score ${M}^{(I)}\in \mathbb{R}^{n\times k}$
 \begin{gather}
{ M }^{ (I) }=\left[ { { H }^{ (f) } }^{ T }{  { H }^{ (b) } }^{ T } \right] \left[ \begin{matrix} Q \\ Q \end{matrix} \right] 
 \end{gather}
 To make sure the attention weight value fall into the range of $[0,1]$, we normalize $M^{(I)}$ to obtain the interaction-attention weight $A^{(I)} = \big ({A}^{(I)}_{tj} \big )_{t=\{1,...,n\},\\j=\{1,...,k\}}$ as follows.
\begin{sequation}
{A}^{(I)}_{tj} =e^{{M}^{(I)}_{tj}}/\sum_{ i=1 }^{n} e^{{M}^{(I)}_{ij}}
\end{sequation}

Similar to self-attention mechanism, the label-aware document representation can be calculated by linear combining the label's context words as ${C}^{(I)}_j=H A^{(I)}_j$,
which can be taken as the representation of the input document along the $j$-th lable. The whole matrix $C^{(I)}\in \mathbb{R}^{2r\times k}$ is the label-aware document represenation under the interaction-attention mechanism.

\vspace{-2mm}
\subsection{Attention Fusion (FA)}
The above ${C}^{(S)}$ and ${C}^{(I)}$ are label-aware document representation. The former focuses on document content, while the latter prefers to the label structure. In order to take advantage of these two parts, an attention fusion strategy is designed here to adaptively extract proper information from these two components and build accurate label-aware document representation. 
More specifically, a fully connected layer is used to transform the input (${C}^{(S)}$ and ${C}^{(I)}$) to weights $\mathbf{\alpha}\in { \mathbb{R} }^{ k\times 1 }$ and $\mathbf{\beta}\in { \mathbb{R} }^{ k\times 1 }$ via
\begin{sequation}
\mathbf{\alpha} =\sigma ({ F }_{ 1 }({ C }^{ (S) })); \quad
\mathbf{\beta} =\sigma ({ F }_{ 2 }({ C }^{ (I) }))
\end{sequation}
where $\sigma$ is sigmoid function to ensure the weights falling into $(0,1)$. Among them, $\alpha_j$ and $\beta_j$ indicates the importances of self-attention and interaction-attetion to final representation along the $j$-th label respectively. Therefore, we normalize them as ${ \alpha  }_{ j }={ \alpha  }_{ j }/({ \alpha  }_{ j }+\beta _{ j})$ and ${ \beta  }_{ j }=1-{ \alpha  }_{ j }$.
With the aid of fusion weights, we can get the final label-aware representation of input document along the $j$-th label 
\begin{sequation}
C_{j}=\alpha_{j} \times { C }_{j}^{ (S) }+\beta_{j} \times { C }_{j}^{ (I) }.
\end{sequation}
The whole matrix $C\in \mathbb{R}^{2r\times k} $ is the final label-aware document represenation.

\subsection{Prediction Layer}
Once having $C\in{ \mathbb{R} }^{ 2r\times k }$, we can build the classifier via a fully connected and output layer. 
The final predictions are obtained by $\hat { y } =\sigma ({ W }_{ o }(f({ W }_{ f }C)))$
where ${ W }_{ f }\in{ \mathbb{R} }^{ r\times 2r }, { W }_{ o }\in{ \mathbb{R} }^{ 1\times r }$, $f$ is the activation function ReLU, and $\sigma$ is adopted to ensure that the output value can be taken as a probability. In this case, 
the binary cross-entropy loss can be used as loss function which has been proved suitable for XMTC tasks~\cite{ref_article6} .
\begin{sequation}
L_{loss}=-\frac { 1 }{ N } \sum _{ i=1 }^{ N }{ \sum _{ j=1 }^{ k }{ [{ y }_{ ij }log(\hat { y } _{ ij })+(1-{ y }_{ ij })log(1-\hat { y } _{ ij })] }  } 
\end{sequation}
where N is the number of training documents. The ground truth $y_{ij}=1$ if the $i$-th document belongs to the $j$-th class, otherwise $y_{ij}=0$.

\section{Experiments}

In this section, we evaluate the proposed \textit{LAHA} on five benchmark datasets by comparing with the state-of-the-art extreme multi-label learning methods in terms of widely used metrics. 

\vspace{-4mm}
\subsection{Datasets}
A series of experiments were carried out on five multi-label datasets with label sizes from 54 to 29,947.The dataset statistics are summarized in Table~\ref{table1}. 

\vspace{-10mm}
\subsection{Methodology}
\subsubsection{Baseline Algorithms}
The proposed \textit{LAHA} is a deep neural network model, thus the recent deep learning-based XMTC methods (\textit{XML-CNN}~\cite{ref_article6} and \textit{AttentionXML}~\cite{ref_article7}) are selected as baselines. 
Meanwhile, the existing powerful 
 \textit{SLEEC}~\cite{ref_article9} (an embedding-based method) and \textit{PfastreXML}~\cite{ref_article10} (a tree-based method) are used as baselines because they obtained the best performance in each type as shown in the Extreme Classification Repository~\footnote{\url{http://manikvarma.org/downloads/XC/XMLRepository.html}.\label{repository}}. 
\vspace{-4mm}
\subsubsection{Parameter Settings}
For all the five datasets, we adopt Glove(300-dimension)~\cite{ref_article12} as word embedding. The number of Bi-LSTM hidden units is set to $r=256$. For the self-attention mechanism, ${d}_{a}=256$. In the prediction layer, ReLU is adopted as non-linear activation function. The whole deep model is trained using Adam with the initial learning rate (0.001) and the batch size (64).

\vspace{-4mm}
\subsubsection{Evaluation Metrics}
In XMTC tasks, 
rank-based evaluation metrics are popular used to evaluate model performance, including Precision at $\tau$ ($P@\tau$) and normalized Discounted Cumulative Gain at $\tau$ ($nDCG@\tau$). Both of them have been widely used in XMTC tasks. They are defined as 
\begin{gather}\label{precK}
P@\tau=\frac { 1 }{ \tau } \sum _{ l\in { r }_{ \tau }(\bm{\mathbf{\hat{y} } }) }{ { \mathbf{y} }_{ l } } ; \quad 
\begin{split}
nDCG@\tau &= \frac{\sum_{l \in r_\tau(\mathbf{\hat{y}})}  \mathbf{y}_l / \log (l+1)}{ \sum_{l=1}^{\min (\tau,||\mathbf{y}||_0)} 1 / \log(l+1)}
\end{split}
\end{gather}
where $\mathbf{y}\in \{ 0,1\} ^{ k}$ is the ground truth label vector of a document and ${ r }_{ \tau }(\bm{\hat { y } })$ is the label indexes of top $\tau$ highest scores of current prediction result.
$\parallel \mathbf{y}\parallel _{ 0 }$ counts the number of relevant labels in the ground truth label vector $\mathbf{y}$. Larger $P@\tau$ and $nDCG@\tau$ indicates better performance.
\vspace{-4mm}
\begin{table}[tbp]
\renewcommand\tabcolsep{4.0pt}
\footnotesize
\centering
\caption{Summary of experimental datasets. $N$ is the number of training documents, $M$ is the number of testing documents, $D$ is the number of features, $L$ is the number of class labels, $\hat { L }$ is the average number of labels per document, $\hat { N } $ is the average number of documents per label.}\label{table1}
\begin{tabular}{lcccccccc}
	\hline
	datsets &  N & M& D& L& $\hat {L} $&$\hat{N} $\\
	\hline
	\textit{AAPD}~\cite{ref_article2} &  54,840 & 1,000 & 69,399 & 54 & 2.41 & 2444.0\\
	\textit{Kan-Shan Cup}~\footnotemark[2] &  2,799,967 & 200,000& 411,721 & 1,999& 2.3  & 3513.1  \\
	\textit{EUR-Lex}~\cite{ref_article3} &  11,585 & 3,865&  171,120&  3,956&  5.3&  15.6\\
	\textit{Amazon-12K}~\cite{ref_article5} &  490,310 & 152,981&  135,895&  12,277&  5.4&  214.5\\
	\textit{Wiki-30K}~\cite{ref_article4} &  12,959 & 5,992&  100,819&  29,947&  18.7&  8.1\\
	\hline
\end{tabular}
\end{table}
\footnotetext[2]{\url{https://biendata.com/competition/zhihu/}}
\subsection{Ablation Test of \textit{LAHA}}
In this section, we firstly
demonstrate the effect of each component on \textit{LAHA}. To reach this goal, we do ablation test for self-attention mechanism (SA), interaction-attention mechanism (IA) and attention fusion mechanism (FA) respectively with two datasets: one sparse dataset \textit{EUR-lex} and one dense dataset \textit{AAPD}.
\vspace{-2mm}
\begin{figure}[h]
	\centering
	\subfigure[ablation test on \textit{EUR-Lex}]{
		\includegraphics[width=1.7in,height=1.5in]{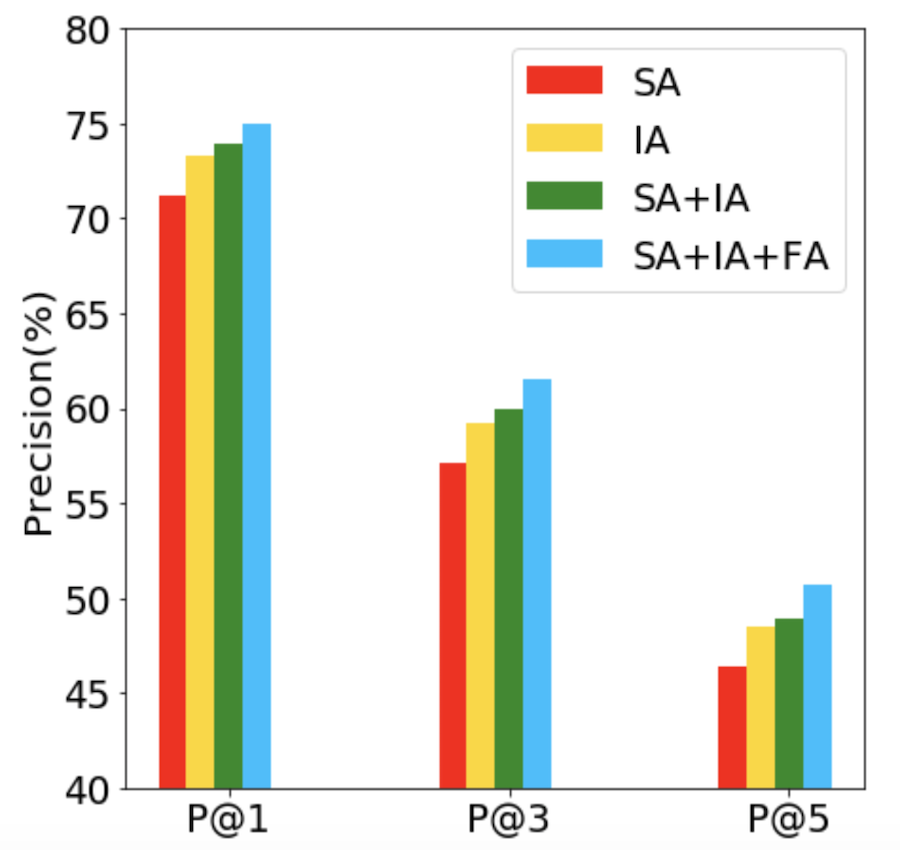}\label{Fig:ablation-a}}
	\subfigure[ablation test on \textit{AAPD}]{
		\includegraphics[width=1.7in,height=1.5in]{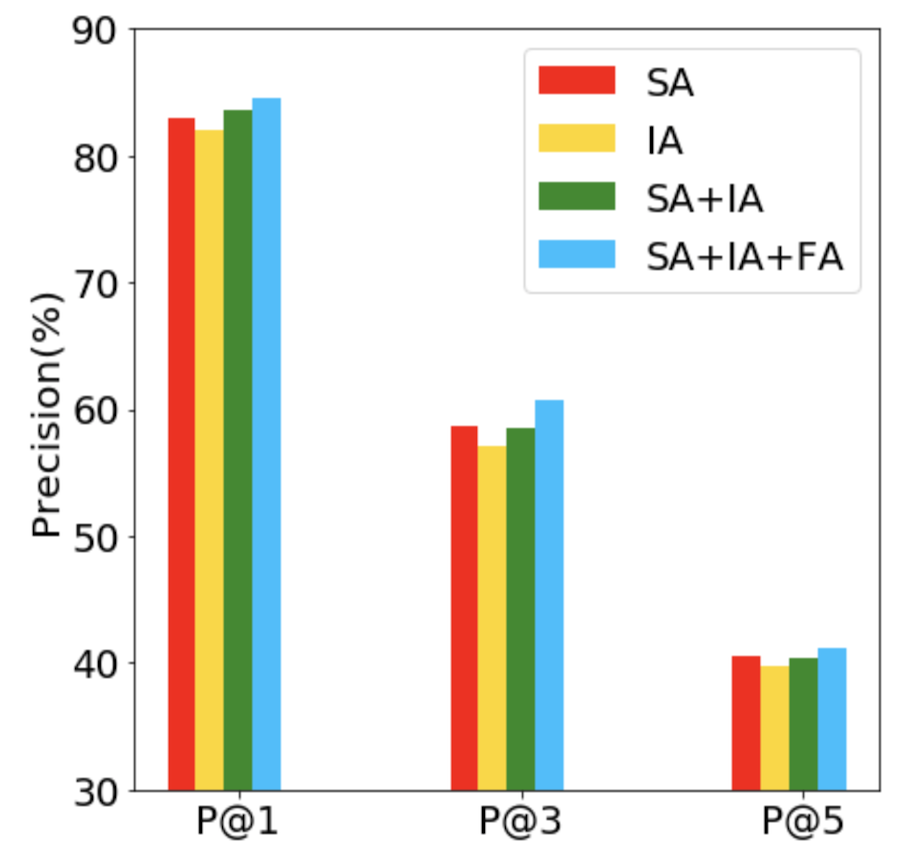}\label{Fig:ablation-b}}
	\setlength{\abovecaptionskip}{0.cm}
	\setlength{\belowcaptionskip}{-0.cm}
	\caption{Ablation test on \textit{EUR-Lex} and \textit{AAPD}. SA=\textit{self-attention}, IA=\textit{interaction-attention}, FA=\textit{attention fusion}, \textit{LAHA}=SA+IA+FA. }\label{Fig:lablation-test} 
\end{figure}

\begin{figure}[h]
	\centering
	\subfigure[\textit{EUR-Lex}]{
		\includegraphics[width=1.7in]{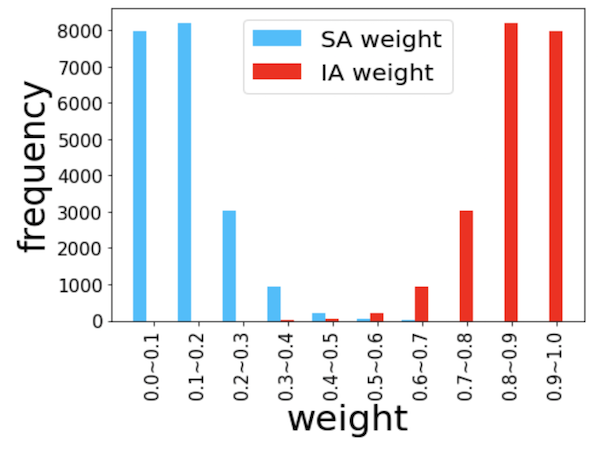}}	
	\subfigure[\textit{AAPD}]{
		\includegraphics[width=1.7in]{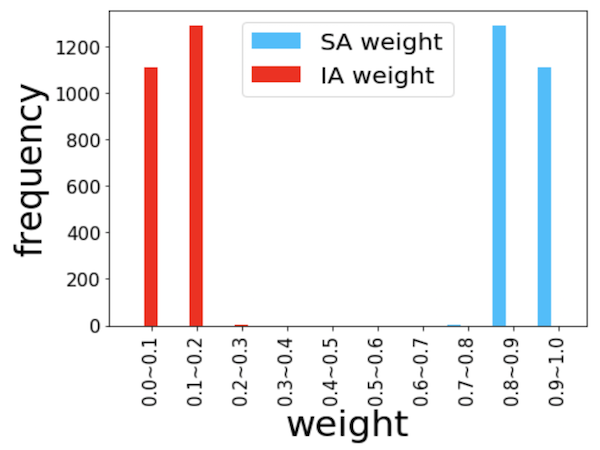}}	
	\setlength{\abovecaptionskip}{0.cm}
	\setlength{\belowcaptionskip}{-0.cm}	
	\caption{Weight distributions for two components on \textit{EUR-Lex} and \textit{AAPD}. $x$-axis indicates the range of weight from 0 to 1 with 0.1 gap. $y$-axis indicates the frequency that the specific range occurs in current label group.} \label{Fig:weight-distribution}
\end{figure}

Fig.\ref{Fig:lablation-test} lists the results on these two datasets in terms of $P@\tau$ ($\tau=\{1,3,5\}$). It can be seen that SA performs well on dense dataset (\textit{AAPD}). However, neither SA nor IA can obtain good result on sparse dataset (\textit{EUR-Lex}). Fortunately, combining SA and IA improves the prediction performance (SA+IA gets better performance than SA and IA). SA prefers to extract the useful content information when constructing the label-aware document representation, but SA ignores the label structure during the learning process. IA implements this by using the label embedding learnt from the label co-exist graph. However, in real application, such graph may contain noisy information (say in dense data). Therefore, coupling with both attention components does really helpful for final performance because they can benefit each other on different datasets.

To adaptively extract proper information to learn the final label-aware document representation, the attention fusion mechanism is introduced in \textit{LAHA}. Fig.\ref{Fig:weight-distribution} lists the distribution of weights on SA and IA. It can be seen that for sparse data (\textit{EUR-Lex}), the interaction-attention plays much more important role than self-attention on learning process, vice verse for dense dataset (\textit{AAPD}). This result further clarifies that IA mechanism can leverage the label structure to improve the prediction performance for sparse data. On the other hand, in \textit{AAPD}, each label has sufficient documents, i.e., SA mechanism can sufficiently capture the label-aware document information and perform well. That is why larger weights are assigned to SA on dense data. 
Similar trend can be found on other datasets, which are omitted due to the page limitation.

\begin{figure}[h]
	\centering
	\subfigure[\textit{	G1($F\le 5$)}]{
		\includegraphics[height=0.8in]{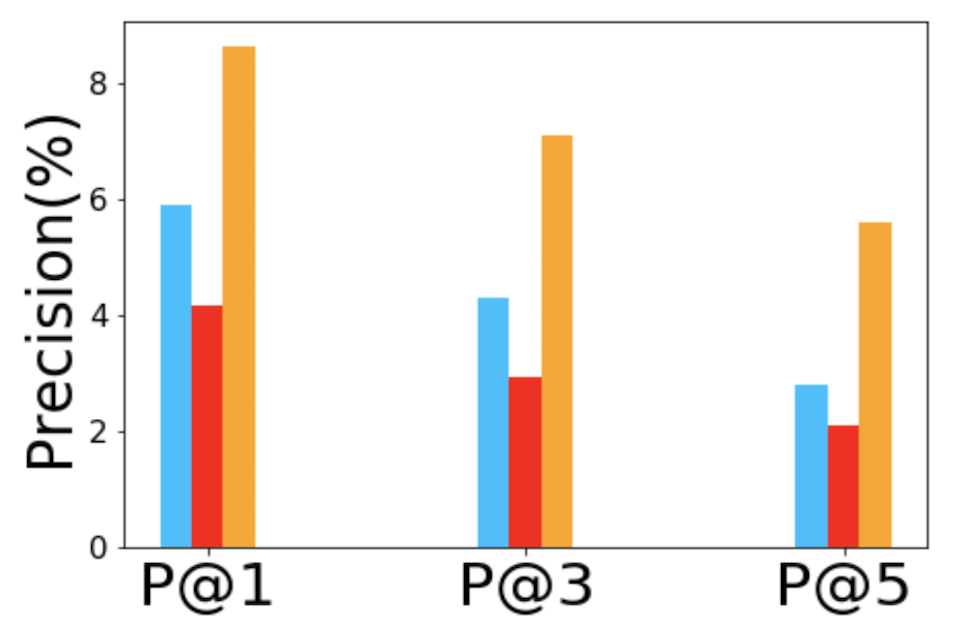}}	
	\subfigure[\textit{G2($5<F\le 50$)}]{
		\includegraphics[height=0.8in]{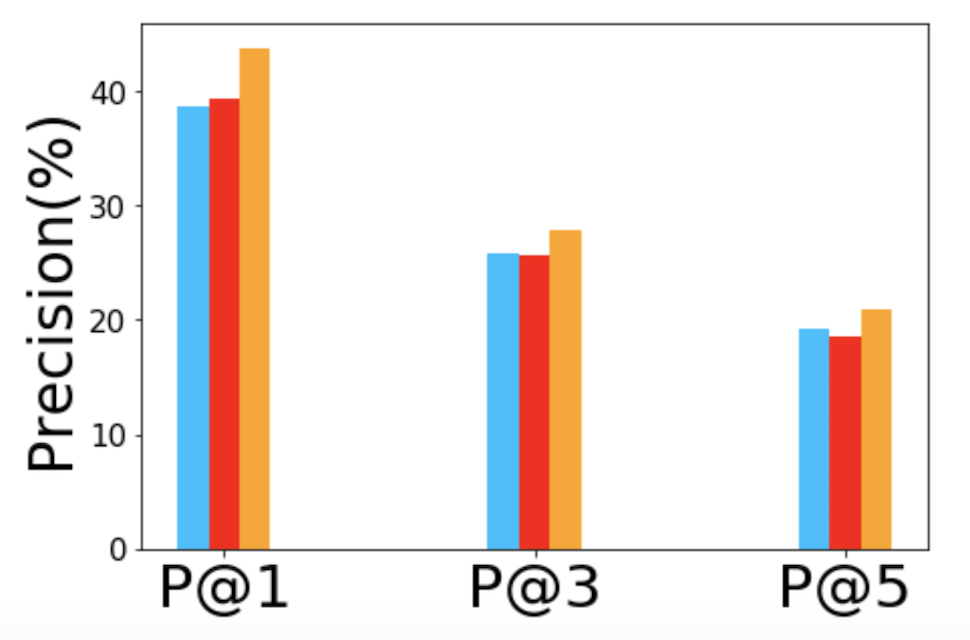}}
	\subfigure[\textit{G3($50<F\le 764$)}]{
		\includegraphics[height=0.8in]{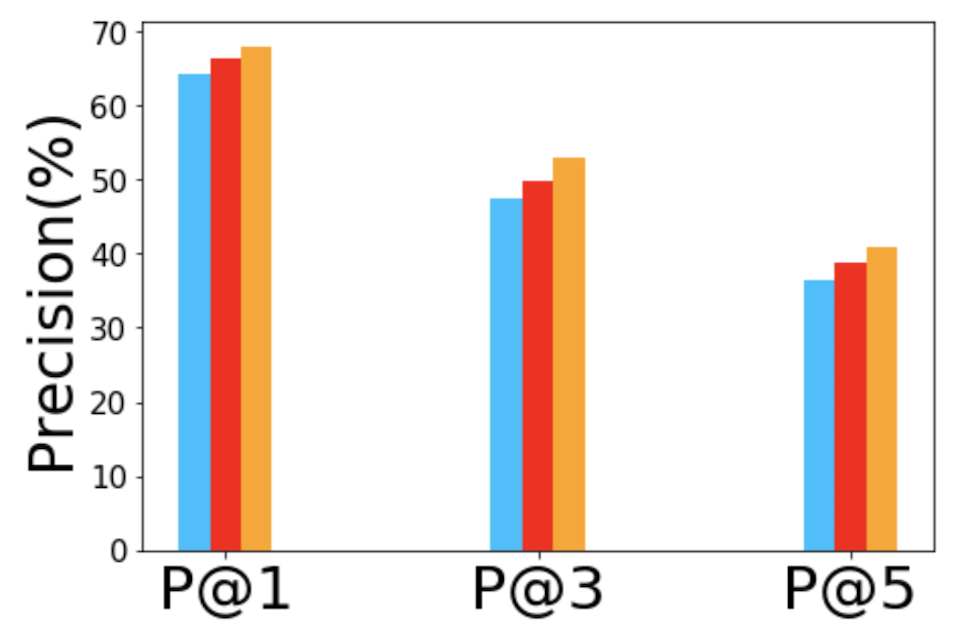}}
	\subfigure{
		\includegraphics[width=3.5in]{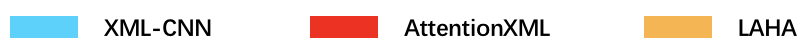}}	
	\setlength{\abovecaptionskip}{0.cm}
	\setlength{\belowcaptionskip}{-0.cm}	
	\caption{Comparing 
		\textit{XML-CNN}, \textit{AttentionXML} and \textit{LAHA} on different label groups of sparse data(\textit{EUR-Lex}) in terms of \textit{P@$\tau$} ($\tau=\{1,3,5\}$). F is frequency of label occurring in training set.} \label{Fig:group-result1}
\end{figure}

\subsection{Comparison with Deep Methods on Sparse Datasets}

In order to explore the effect of \textit{LAHA} on sparse datasets, we further divide labels into three groups according to their occurring frequencies. Fig.\ref{Fig:group-result1} shows the prediction performance obtained by three deep methods. Obviously, label prediction in \textit{G1} is much harder than in other two groups due to the lack of training documents.
All methods become better from \textit{G1} to \textit{G3}, which is reasonable since \textit{G3} contains more training documents than \textit{G1}.  \textit{LAHA} has an overall improvement for all groups compared with two baselines. This result further demonstrates the superiority of the proposed hybrid attention mechanism on XMTC with large-scale tail labels. Similar phenomena can be found on other sparse datasets, which are omitted due to the page limitation.
\hspace{-2mm}
{\small 
\begin{figure}[h]
	\centering
	\subfigure[The words with largest label-aware attention weights output by \textit{AttentionXML} (word$\rightarrow $\{labels\}) are: \textit{autism$\rightarrow $\{autism\}; cure others$\rightarrow $\{disease\}; disorder$\rightarrow $\{abnormal\}; social communication, approach others$\rightarrow $\{social norm\}; children$\rightarrow $\{children, childhood\}}.]{
		\includegraphics[height=1in,width=4.5in]{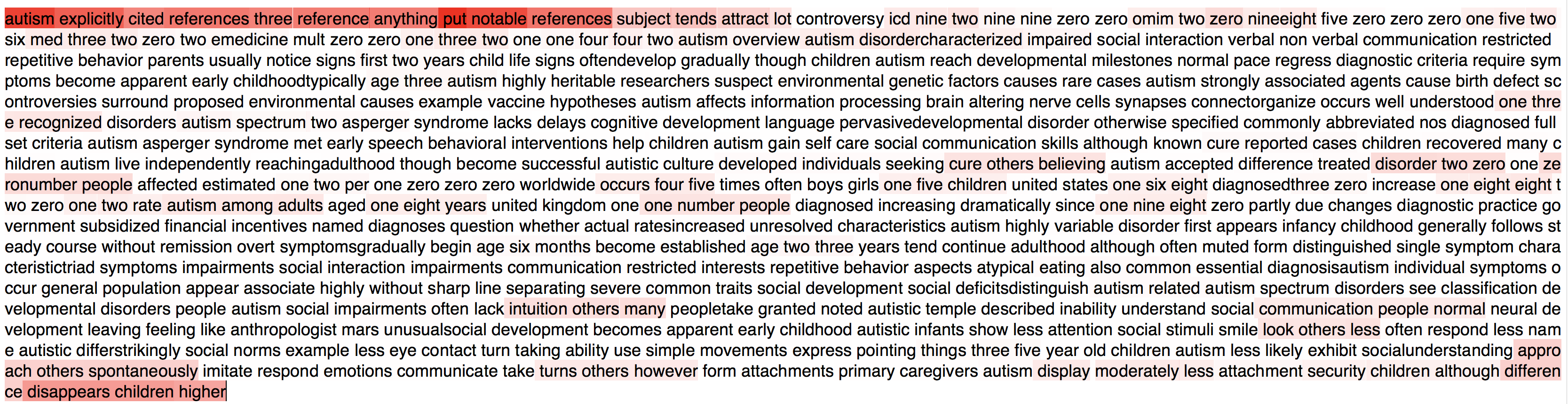}\label{Fig:dataset-a}}
	\subfigure[The words with largest label-aware attention weights output by \textit{LAHA} (word$\rightarrow $\{labels\}) are: \textit{autism$\rightarrow $\{autism\}; disorder$\rightarrow $\{abnormal\}; child life, infancy childhood, children$\rightarrow $\{children, childhood\};  diagnostic, genetic factor, lack intuition$\rightarrow $\{disease\}; synapses connect organize$\rightarrow $\{neurology\}; asperger syndrome$\rightarrow $\{asperger\}; social communication$\rightarrow $\{social norm\}; security$\rightarrow $\{disease, health\}}.]{
		\includegraphics[height=1in,width=4.5in]{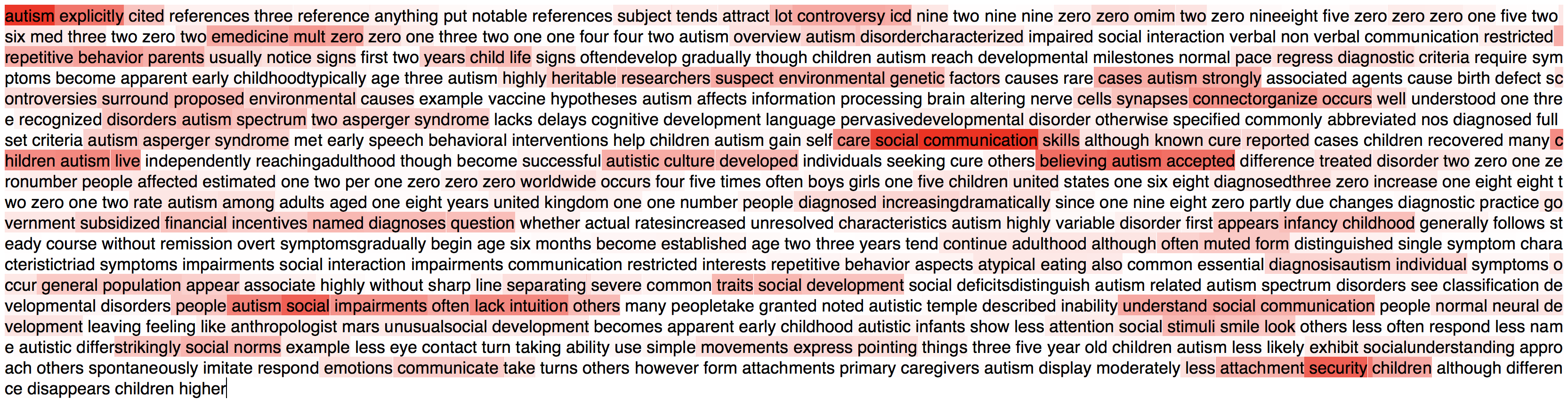}\label{Fig:dataset-b}}
	\setlength{\abovecaptionskip}{0.cm}
	\setlength{\belowcaptionskip}{-0.cm}
	\caption{Heat map of label-aware attention weights obtained by (a) \textit{AttentionXML} and (b) \textit{LAHA} on an example document from \textit{Wiki30K}.}\label{Fig:heat-map} 
\end{figure}
}

To further investigate the attention-based methods, we visualize the attention weights on the original document using heat map, as shown in Fig.\ref{Fig:heat-map}. This example document belongs to 28 labels named as \textit{autism, children, childhood, disease, asperger, social norm, health, neurology, abnormal} and etc. From the attention weights, we can see that \textit{AttentionXML} only captures few key words for few related labels. As expected, \textit{LAHA} focuses on the related information as much as possible due to the capacity making full use of label structure and document content. 

\vspace{-4mm}
\subsection{Comparison Results and Discussion}
In this section, the proposed \textit{LAHA} is evaluated on five benchmark datasets by comparing with four baselines in terms of $P@\tau$ and $nDCG@\tau$ ($\tau=\{1,3,5\}$). Table \ref{table2} shows the averaged performance of all test documents. According to the formula (\ref{precK}), we know $P@1=nDCG@1$, thus only $nDCG@3$ and $nDCG@5$ are listed. In each line, the best result is marked in bold, and the second best is underlined.

\begin{table}[t]
	\centering
	\small
	\caption{Comparing \textit{LAHA} with four baselines in terms of various metrics on five benchmark datasets.}\label{table2}
	\begin{tabular}{l|c|c|c|c|c|c}
		\hline
		Datasets  &Metric &\textit{SLEEC} & \textit{PfastreXML} &\textit{XML-CNN} &\textit{AttentionXML}     & \textit{LAHA}	  \\ \hline
		\multirow{5}{*}{\textit{AAPD}}	 &$P@1$ & 81.96\%& 82.35\% &76.25\% & \underline{83.02\%} &  \textbf{84.48\%} 	  \\ 
		& $P@3$ &57.48\% & 58.01\% & 54.34\%& \underline{58.72\%} &  \textbf{60.72\%} 	 \\
		& $P@5$ & 38.99\%& 40.13\% & 37.84\%& \underline{40.56\%} &  \textbf{41.19\%} 	  \\ \cline{2-7}
		& $nDCG@3$ & 77.65\%& \underline{78.26\%} & 72.01\%& 78.01\% & \textbf{80.11\%} 	 \\
		& $nDCG@5$ & 81.59\%& 82.03\% &76.40\% & \underline{82.31\%} & \textbf{83.70\%} 	  \\ \hline \hline
		\multirow{5}{*}{\textit{Kan-Shan Cup}}	 &$P@1$ & 51.41\%& 52.29\% &49.68\%& \underline{53.69\%} 	& \textbf{54.38\%} 	  \\ 
		& $P@3$ & 32.81\%& 32.99\% & 32.27\%& \underline{34.10\%} 	& \textbf{34.60\%} 	 \\
		& $P@5$ & 24.29\%& 24.58\% &24.17\% & \underline{25.16\%} & 	 \textbf{25.88\%} 	  \\ \cline{2-7}
		& $nDCG@3$ & 49.32\% & 49.96\% & 46.65\%& \underline{51.03\%} & \textbf{51.70\%} 	 \\
		& $nDCG@5$ & 49.74\%& 50.11\% &49.60\% & \underline{53.96\%} &  \textbf{54.65\%} 	  \\ \hline \hline
		\multirow{5}{*}{\textit{EUR-Lex}}	 &$P@1$ &\textbf{75.18\%} &  73.03\%&70.94\% & 71.89\% 	&  \underline{74.95\%}	  \\ 
		& $P@3$ & \textbf{61.67\%}& 60.39\% &56.02\% & 57.74\% &  \underline{61.48\%}	  \\
		& $P@5$ & \underline{50.23\%}& 49.69\% &45.36\% & 47.35\% &  \textbf{50.71\%} 	 \\ \cline{2-7}
		& $nDCG@3$ & \underline{63.79\%}& 62.51\% &59.68\% & 61.29\% &  \textbf{64.89\%}	  \\
		& $nDCG@5$ & \underline{58.03\%}& 57.72\% &53.82\% & 56.71\% &  \textbf{59.28\%} 	 \\ \hline \hline
		\multirow{5}{*}{\textit{Amazon-12K}}	 &$P@1$ & 93.49\%& \underline{93.95\%} &93.15\% & 93.75\% &  \textbf{94.87\%} 	  \\ 
& $P@3$ & 78.01\%& 78.33\% & 76.11\%& \underline{78.36\%} &  \textbf{79.16\%}	 \\
& $P@5$ & 62.09\%& \underline{62.77\%} &60.51\% & 62.14\% &  \textbf{63.16\%} 	  \\ \cline{2-7}
& $nDCG@3$ & 86.89\%& \underline{88.41\%}& 86.75\%& 87.62\% &  \textbf{89.13\%}	 \\
& $nDCG@5$ & 84.53\%& \underline{86.23\%} &84.01\% & 86.06\% &  \textbf{87.57\%} 	  \\ \hline	\hline
		\multirow{5}{*}{\textit{Wiki-30K}}	 &$P@1$ & \textbf{85.26\%}& 82.81\% & 82.90\%& 81.98\% &  \underline{84.18\%} 	  \\ 
		& $P@3$ & \textbf{73.91\%}& 68.48\% & 67.46\%& 67.27\% & \underline{73.14\%} 	 \\
		& $P@5$ & \underline{62.55\%}& 59.93\% &57.09\% & 56.43\% &  \textbf{62.87\%} 	  \\ \cline{2-7}
		& $nDCG@3$ & \textbf{76.01\%}& 72.15\% & 71.04\%& 70.77\% & \underline{75.64\%} 	 \\
		& $nDCG@5$ & \textbf{68.27\%}& 63.83\% & 62.92\%& 62.35\% & \underline{67.82\%} 	  \\ \hline\hline
		\multicolumn{2}{c|}{\textit{Win times}} & 6 & 0&0&0 & 19\\
		\hline	
	\end{tabular}
\end{table}

From Table \ref{table2}, we can make a number of observations about these results. Firstly, \textit{LAHA} outperforms the traditional powerful embedding-based and tree-based methods in most cases, while slightly underperforms the embedding-based method \textit{SLEEC} on \textit{EUR-Lex} and \textit{Wiki-30K}. From Table \ref{table1}, we can see there are only 11,585 and 12,959 training documents in these two datasets, in this case, the deep model may be not sufficiently trained.
Second, \textit{LAHA} is consistently superior to the state-of-the-art deep XMTC methods. The main reason is that \textit{LAHA} has ability to sufficiently determine the label-aware document representation while \textit{XML-CNN} does not. Even though \textit{AttentionXML} tries to find the relation between each pair of document and label, it only focuses on document content, which will degrade its performance on tail labels due to lack of information. Fortunately, \textit{LAHA} addresses this issue by simultaneously considering label structure via a hybrid attention mechanism.


\vspace{-4mm}
\section{Conclusions and Future Work}
In this paper, a new XMTC method, \textit{LAHA}, is proposed. \textit{LAHA} utilizes self-attention and interaction-attention to extract the semantic relation between words and labels, and an attenton fusion to construct the label-aware document representation. Extensive experiments on five benchmark datasets prove the superiority of \textit{LAHA} by comparing with the state-of-the-art XMTC methods.
In a nutshell, the novelty of \textit{LAHA} lies in its providing a label-aware document representation that captures both document content and label structure, and has better discriminative ability than baselines.
In real applications, more contents can be collected such as label content, which is proved to be helpful in XMTC~\cite{ref_article8}. We therefore plan to extend the current model with such information.

%
%
%
%

\end{document}